\documentclass[runningheads]{llncs}

\usepackage{eccv}

\usepackage{eccvabbrv}

\usepackage{graphicx}
\usepackage{booktabs}
\usepackage{lipsum}

\usepackage[accsupp]{axessibility}  

\usepackage{hyperref}

\usepackage{orcidlink}
\usepackage{xcolor}
\definecolor{projectpink}{RGB}{220, 70, 150}

\usepackage{multirow}
\usepackage{marvosym}

\begin{document}

\title{CameraAnything: Refilming Videos with Arbitrary Camera Control} 

\titlerunning{CameraAnything: Refilming Videos with Arbitrary Camera Control}
\author{
Yixuan Li$^{1,2,*}$, 
Yanhong Zeng$^{2,3,*,\dagger}$, 
Ka Leong Cheng$^{2}$, 
Jiayi Zhu$^{2}$, 
Hanlin Wang$^{2,5}$, 
Wen Wang$^{2,4}$, 
Yihao Meng$^{2,5}$, 
Hao Ouyang$^{2}$, 
Qiuyu Wang$^{2}$, 
Yue Yu$^{2,5}$, ZiDong Wang$^{1}$, Yiyuan Zhang$^{1}$, Yujun Shen$^{2}$, Dahua Lin$^{1}$
}
\institute{
The Chinese University of Hong Kong, Hong Kong, China\\
\email{liyixxxuan@gmail.com}\and
Ant Group, Hangzhou, China\\
\email{zengyh1900@gmail.com}\and
Tsinghua University, Beijing, China\and
Zhejiang University, Hangzhou, China\and
The Hong Kong University of Science and Technology, Hong Kong, China
}

\authorrunning{Yixuan Li et al.}


\maketitle
\begin{center}
\footnotesize
$^{*}$Equal contribution.\quad $^{\dagger}$Project lead.
\end{center}
\vspace{0.5em}
{\centering\textcolor{projectpink}{\url{https://yixuanli98.github.io/cameraanything/}}\par}
\begin{center}
    \centering
    \includegraphics[width=1.0\textwidth]{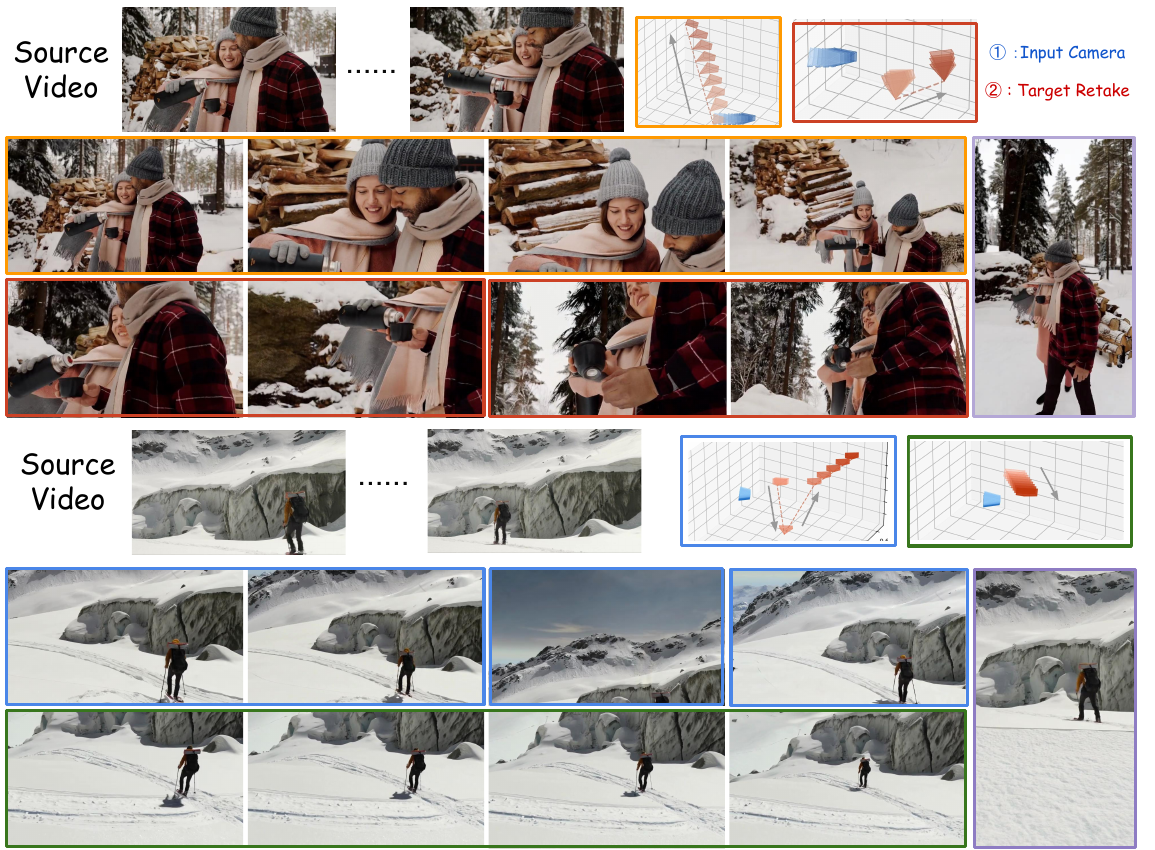}
    \label{fig:teaser}
    \captionof{figure}{
    We propose \textbf{CameraAnything}, the first unified video editing framework that enables comprehensive camera control, including novel trajectory manipulation, multi-shot transitions, resolution adaptation, and focal length adjustment in a single framework, demonstrating strong potential for cinema-level video editing.}
\end{center}

\begin{abstract}
We introduce \textbf{CameraAnything}, the first unified framework for camera controlled video editing that enables joint control of both intrinsic and extrinsic camera parameters. Existing approaches either rely on expensive 3D reconstruction to achieve full camera functionality or restrict editing to extrinsic parameter manipulation. Moreover, the coupled influence of intrinsic and extrinsic parameters on video appearance makes disentangled modeling particularly challenging.
To address this, we adopt per-pixel Pl\"ucker ray injection alongside resolution-aware 3D RoPE in self-attention, building both camera conditioning and spatial positional encoding on the target latent to jointly control camera position, focal length, and native resolution editing without cropping or outpainting. To overcome the scarcity of paired training data, we further develop a scalable synthetic pipeline that constructs diverse dynamic scenes through structured multi-camera recording and generates synchronized videos with varied camera configurations.
With a tailored orthogonal training strategy, CameraAnything enables expressive video reshooting with arbitrary viewpoint control, focal length adjustment, resolution adaptation, and multi-shot transitions within a single generation process, offering strong practical value for cinematic video editing and cross-platform content adaptation in video production.

\end{abstract}

\section{Introduction}

Controlling the camera is fundamental to visual storytelling. Recent advances in generative video models enable controllable camera movements, allowing users to reshoot existing videos with novel camera motions \cite{bai2025recammaster,park2025redirector,fu2026plenoptic}. Such capabilities promise to democratize cinematic video production by enabling complex camera effects without requiring professional equipment or multi-camera setups.

Expressive camera language arises from the joint manipulation of \textbf{extrinsic} parameters (camera position and orientation) and \textbf{intrinsic} parameters (e.g., focal length and resolution), which determine scene projection, framing, and viewer attention. Despite promising progress, state-of-the-art camera-controlled video editing methods support only extrinsic pose control and lack intrinsic parameter manipulation such as focal length, limiting their ability to reproduce expressive camera language \cite{bai2025recammaster,park2025redirector,fu2026plenoptic}. Moreover, many methods assume fixed resolutions or aspect ratios, restricting applicability across diverse display platforms.

In this paper, we present \textbf{CameraAnything}, the first unified framework for camera-controlled video editing that supports expressive camera language. As illustrated in Fig.~\ref{fig:teaser}, it enables four forms of camera control: viewpoint manipulation, focal length adjustment, resolution adaptation, and multi-shot transitions within a single generation process. Given an input video and target camera parameters, CameraAnything conditions a video diffusion transformer on camera-aware geometric representations to generate videos that follow the specified configuration. A key challenge is modeling camera parameters that jointly affect scene appearance, as intrinsic and extrinsic parameters are often tightly coupled in cinematic operations. For example, the Hitchcock dolly zoom simultaneously changes focal length and camera position to maintain the subject's apparent size while dramatically altering background perspective \cite{adobe_dolly_zoom}.

To address this problem, we adopt per-pixel Plücker ray injection alongside resolution-aware 3D RoPE in self-attention, building both camera conditioning and spatial positional encoding on the target latent. We represent camera geometry using token-wise Plücker rays~\cite{sitzmann2021light}, which encode the 3D viewing direction and moment of each pixel ray, uniquely representing the camera ray associated with each image token. This formulation naturally captures both intrinsic and extrinsic camera parameters. Following Rotary Camera Encoding (RoCE)~\cite{park2025redirector}, we inject these embeddings into Rotary Position Encoding (RoPE) within the transformer attention layers~\cite{su2024roformer}, augmenting the positional encoding used for query--key interactions. For native resolution editing, we directly recompute the spatial coordinates of 3D RoPE from the target-resolution latent grid, so token positional encoding adapts to the desired output size. By grounding visual tokens in camera geometry while preserving RoPE's positional generalization, the model learns geometry-consistent representations that remain stable under diverse camera configurations. In practice, we evaluate several representation and injection variants through ablation and adopt the configuration above as the most effective for unified camera control.
Training such a model requires paired videos with controlled camera variations, which are difficult to collect in the real world. We build a scalable synthetic data pipeline based on structured multi-camera recording, rendering synchronized clips per scene with diverse extrinsic trajectories, focal changes, and aspect-ratio variations. We further employ an orthogonal training strategy that independently samples extrinsic, focal, and resolution edits, enabling both single-dimension control and compound camera instructions.

Extensive experiments demonstrate that CameraAnything achieves state-of-the-art performance across multiple camera-controlled video editing tasks, achieving expressive camera language and facilitating practical video adaptation across different display formats.
Our contributions are summarized as follows:
\begin{itemize}
\item We introduce \textbf{CameraAnything}, a unified framework for camera-controlled video editing that jointly manipulates viewpoint, focal length, resolution, and multi-shot transitions within a single generation process. To realize these functions, we inject per-pixel Plücker camera embeddings alongside resolution-aware 3D RoPE on the target latent, as selected through ablation over alternative representation and injection designs.

\item We build a scalable synthetic data pipeline using structured multi-camera recording to generate synchronized training videos with diverse and controllable camera configurations, and design an orthogonal training strategy that independently samples extrinsic, focal, and resolution edits to support both single-dimension control and compound camera instructions.

\item Extensive experiments demonstrate state-of-the-art performance on camera-controlled video editing tasks and highlight the framework's potential for cinematic video production and cross-platform content adaptation.
\end{itemize}

\section{Related works}

\paragraph{Camera-controlled video generation.}
Camera-controlled video generation has attracted growing research interest for its potential to enable film-level narrative storytelling and democratize individual film production. A central challenge in this area is how to effectively represent and inject camera parameters into video generation models.
Early approaches adopt explicit camera parameterizations such as Plücker embeddings or extrinsic matrices. CameraCtrl \cite{he2024cameractrl} introduces a camera encoder that maps Plücker embeddings \cite{sitzmann2021light} into multi-scale features, which are integrated into all temporal attention layers of a U-Net–based video model. CameraCtrl-II \cite{he2025cameractrl} extends this by curating a dataset with extensive dynamics and camera annotations, injecting camera representations via element-wise addition before the first DiT layer to improve viewpoint range and motion diversity. SynCamMaster \cite{baisyncammaster} directly uses camera extrinsic matrices and trains on Unreal Engine–rendered multi-camera videos with a hybrid training scheme to support multi-camera generation. ShotDirector \cite{wu2025shotdirector} combines both representations through a dual-branch design with an extrinsic branch and a Plücker branch, enabling multi-shot generation with flexible shot transitions.
More recent works explore tighter integration of 3D geometry with visual tokens in Transformer architectures, including token-level raymap encodings \cite{gao2024cat3d,sitzmann2021light}, attention-level relative pose encodings \cite{miyatogta,kong2024eschernet}, and relative projective positional encoding (PRoPE) with full frustum representations \cite{li2025cameras,hunyuanworld2025tencent,zhang2025unified}.

\paragraph{Camera-controlled video editing.}
While video editing has been extensively studied for tasks such as style transfer, local editing, and inpainting \cite{vace,zhong2025outdreamer}, camera-controlled video editing, which aims to re-render an existing video from novel viewpoints, remains comparatively underexplored.
One line of work reconstructs explicit 3D representations (e.g., point clouds, meshes) from reference videos and then re-renders them from novel camera trajectories, often followed by quality enhancement to handle artifacts and enable additional controllability such as focal length and illumination adjustments \cite{Yu_2025_ICCV,chen2025postcam,hu2025ex,seo2025vid,song2025worldforge,hong2025inversecrafter,liu2025light}. However, these methods are constrained by the fidelity of 3D reconstruction, particularly for complex or dynamic scenes.
Another line of work directly trains generative models to re-shoot videos with new camera trajectories. ReCamMaster \cite{bai2025recammaster}, a pioneering effort in this direction, trains on large-scale multi-camera synthetic synchronized videos and uses camera extrinsic matrices to condition generation. ReDirector \cite{park2025redirector} proposes RoCE, which encodes camera information through relative positional embeddings in the spatial component of 3D-RoPE within self-attention layers. PlenopticDreamer \cite{fu2026plenoptic} synchronizes generative hallucinations across multi-view video re-rendering by maintaining spatio-temporal memory through autoregressive generation with camera-guided video retrieval.
Beyond static viewpoint changes, recent works further explore joint camera control with additional conditions, such as world time \cite{wang2025bullettime,huang2025spacetimepilot}, professional photographic parameters including bokeh, exposure, and color tone \cite{yuan2025generative,sun2025generative,yang2025cameramaster}, or replicating camera movements from reference videos without requiring explicit camera parameters \cite{luo2025camclonemaster}. However, existing methods are generally limited to reshooting videos with smooth, continuous extrinsic camera pose transitions along a single trajectory, without supporting intrinsic camera parameter adjustments.
In contrast, our work presents the first unified framework that enables both multi-shot camera pose changes within a single trajectory and intrinsic parameter adjustments including focal length and resolution, offering significant practical value for adapting content across different display devices, such as portrait mode for mobile screens and horizontal formats for television broadcasts.

\section{Dataset}
\label{sec:dataset}

 To support our proposed framework—the first to enable re-shooting single-shot videos with novel trajectories, multi-shot editing, and intrinsic control—a specialized dataset is required. While ReCamMaster~\cite{bai2025recammaster} provides a foundation for modifying trajectories, it is limited by static camera intrinsics and continuous trajectories, failing to support re-focal effects or the transition from a single long take to a cinematic multi-shot edit. Consequently, we develop a high-fidelity synthetic dataset using Unreal Engine 5~\cite{ue}, enabling joint control over camera extrinsics ($R, T$) and intrinsics, including focal length $f$ and resolution ($H, W$).

\begin{figure*}[!htbp]
    \centering
    \includegraphics[width=\linewidth]{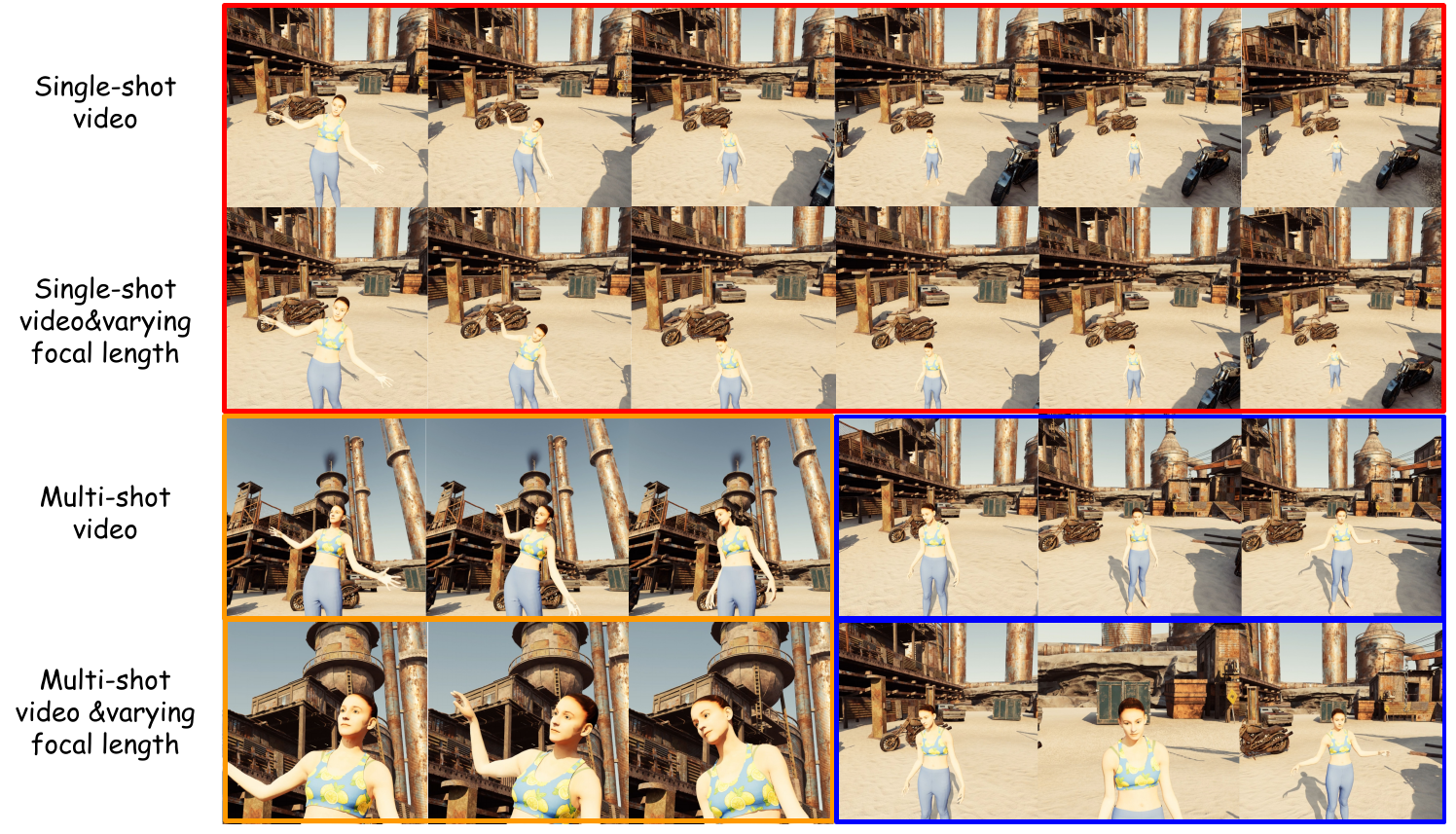}
    \caption{Dataset Example. The red box represents single-shot settings with consistent camera extrinsics, including both fixed and varying focal lengths. The orange and blue boxes each represent a different shot in multi-shot sequences.}
    \label{fig:dataset_example}
\end{figure*}

We construct diverse 3D environments populated with animated human characters to simulate real-world dynamic scenes. To enable the model to capture the spatial correlations between alternative viewpoints and the temporal continuity across cinematic cuts, we employ a structured recording protocol. For each scene, we render 20 synchronized video clips, strictly aligned in the time dimension but varying in camera and editing configurations. The 20 clips per scene are organized into three distinct groups to facilitate multi-task learning: 
\begin{itemize}
\item Group A (Base Views, Clips 1-5): These clips are rendered as continuous, single-shot trajectories with a standard fixed focal length. They serve as the raw input "anchors" providing the base visual and motion information.
\item Group B (Multi-shot Editing Variants, Clips 6-10): These clips replace smooth trajectories with \textbf{1--3} instantaneous camera cuts per video, where the camera undergoes abrupt position and orientation changes to simulate cinematic shot transitions. By pairing Group-A single-shot sources with these multi-shot targets, the model learns to transform a continuous long take into an edited sequence with distinct shots.
\item Group C (Re-focal Variants, Clips 11–20): Unlike prior frameworks that keep camera intrinsics fixed, we construct this group to enable re-focal capability. These clips reuse the exact extrinsic trajectories from Groups A and B but are rendered with different focal lengths. By pairing clips with identical camera motion yet varying fields of view (FoV), the model learns to disentangle geometric camera motion from optical zooming effects, enabling more flexible focal length control during generation.
\end{itemize}

Beyond focal length, we introduce a strategy to achieve native resolution and aspect ratio editing, which we treat as critical components of camera intrinsics. We render all raw footage in a high-resolution square format ($1:1$) to serve as a versatile data source. During training, we perform dynamic, parameterized center cropping on these square frames to generate training samples. This approach allows the model to learn how to re-compose the scene for various output formats (e.g., 16:9, 9:16) and resolutions while maintaining visual fidelity—a capability entirely absent in previous fixed-intrinsic frameworks. We provide some cases in Fig.~\ref{fig:dataset_example}. Additional dataset details are provided in Appendix A.

\begin{figure*}[t]
    \centering
    \includegraphics[width=\linewidth]{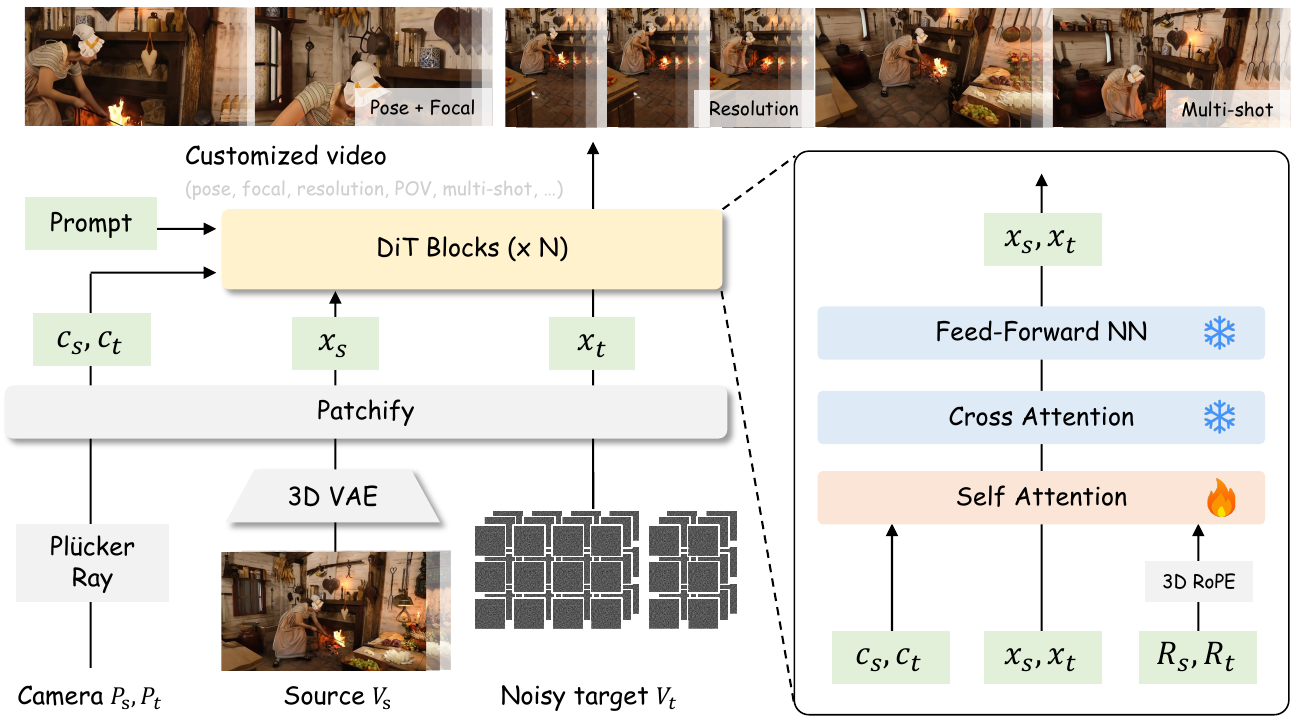}
    \caption{\textbf{Overview of CameraAnything.} CameraAnything is built upon Wan2.1-T2V-1.3B \cite{wan2025wan}, which includes a 3D VAE for video latent encoding and a diffusion Transformer composed of multiple DiT blocks. Given a source video together with its camera parameters and a target camera specification, CameraAnything represents cameras using Plücker rays to provide accurate per-token geometric modeling, which is injected into the self-attention modules. With this design, CameraAnything can generate videos under diverse camera edits, including novel camera poses, focal length adjustments, resolution adaptation, and multi-cut shot transitions, while also supporting flexible combinations of these controls within a single generation process.}
    \label{fig:overview}
\end{figure*}

\section{CameraAnything}
\label{sec:method}
\subsection{Preliminary}

\paragraph{Video Diffusion Transformer.}
As illustrated in \cref{fig:overview}, we build on Wan2.1-T2V-1.3B~\cite{wan2025wan}, a latent video diffusion model that encodes frames into a compact latent space using a 3D Variational Autoencoder (VAE) and denoises them with a Diffusion Transformer (DiT). The 3D-VAE compresses a video $V \in \mathbb{R}^{F \times H \times W \times 3}$ into a latent $z \in \mathbb{R}^{F' \times H' \times W' \times C}$, with temporal and spatial strides of $4$ and $8$ respectively. The DiT processes $N = F' \times H' \times W'$ spatiotemporal tokens obtained by spatially patchifying the latent. Each DiT block applies timestep-conditioned adaptive layer normalization (AdaLN) before self-attention and feed-forward layers, while 3D Rotary Positional Embeddings (RoPE) encode the spatiotemporal coordinates $(f, h, w)$ of each token in the attention computation.

For video-to-video generation, we follow ReCamMaster~\cite{bai2025recammaster} and concatenate the source video tokens $x_s$ and target video tokens $x_t$ along the frame dimension, forming a joint sequence of length $2N$ as input to the DiT. This \emph{frame-dimension conditioning} enables direct spatiotemporal attention between source and target tokens through every self-attention layer, yielding stronger content consistency than channel-dimension or view-dimension alternatives~\cite{bai2025recammaster}.

\paragraph{Camera Parameterization.}
A camera at frame $i$ is described by its camera-to-world (c2w) extrinsic matrix $\mathbf{P}_i = [\mathbf{R}_i \mid \mathbf{t}_i] \in \mathbb{R}^{3 \times 4}$ and intrinsic matrix $\mathbf{K}_i \in \mathbb{R}^{3 \times 3}$. Following standard pinhole optics, we derive $\mathbf{K}_i$ from the horizontal field-of-view:
\begin{equation}
  f_x = \frac{W}{2\tan(\mathrm{hfov}/2)},\quad
  \mathbf{K}_i = \begin{pmatrix} f_x & 0 & W/2 \\ 0 & f_x & H/2 \\ 0 & 0 & 1 \end{pmatrix}.
\end{equation}
All camera poses are expressed relative to the first frame of the source video,
\begin{equation}
  \hat{\mathbf{P}}_i = \mathbf{P}^{-1}_{\mathrm{src},0} \cdot \mathbf{P}_i,
\end{equation}
so that $\hat{\mathbf{P}}_0 = \mathbf{I}$ and subsequent poses encode viewpoint changes in a coordinate-system-agnostic manner.

\begin{figure*}[t]
    \centering
    \includegraphics[width=\linewidth]{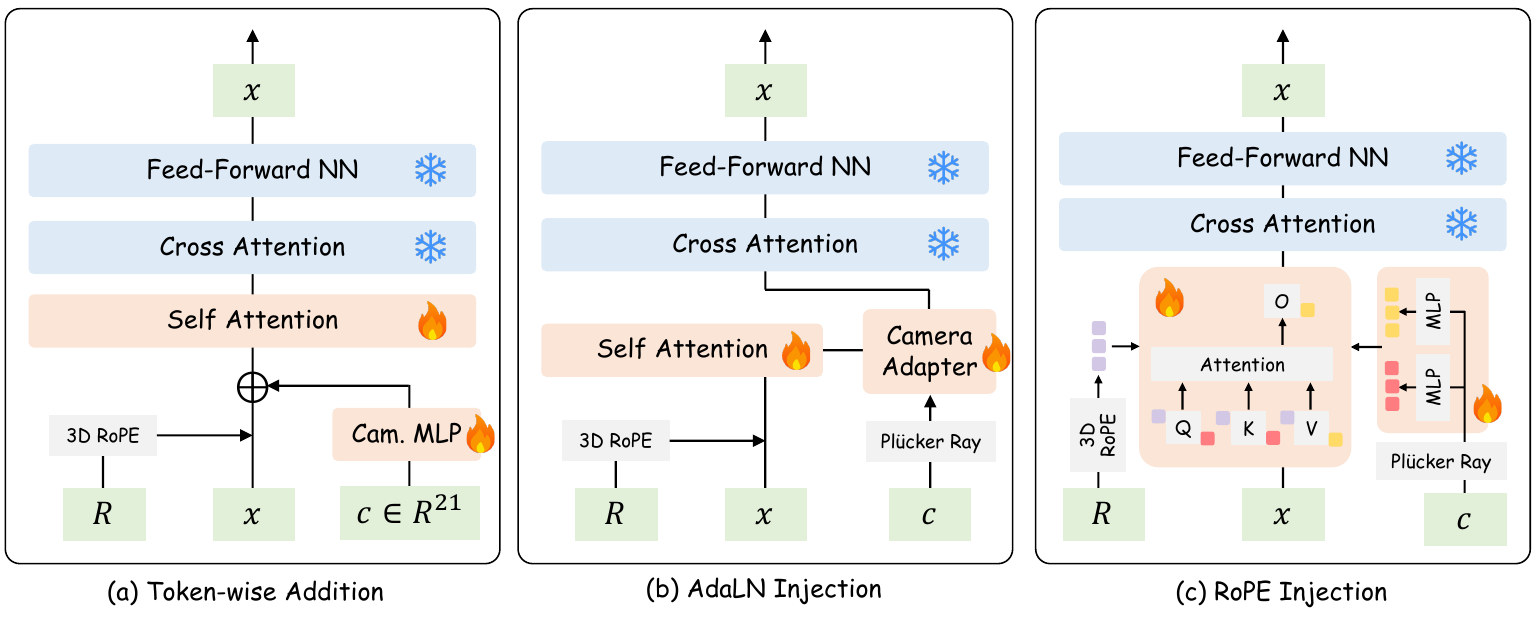}
    \caption{Illustration of three camera representation and injection mechanisms. \textbf{(a) Token-wise Addition}: $\mathbf{P}_i \in \mathbb{R}^{3 \times 4}$ and $\mathbf{K}_i \in \mathbb{R}^{3 \times 3}$ are flattened and concatenated into $\mathbf{c}_i \in \mathbb{R}^{21}$, then linearly encoded and added to all tokens. \textbf{(b) AdaLN Injection}: per-pixel camera embeddings based on the Plücker ray representation are encoded by a learnable linear layer to generate scale and shift parameters that modulate the Layer Normalization layers. \textbf{(c) RoPE Injection}: embedded Plücker rays are mapped to phase shifts that act as positional encodings, applied in parallel with the original 3D RoPE during self-attention computation. Red flames indicate trainable components, while snowflakes denote frozen weights.}
    \label{fig:camera}
\end{figure*}

\subsection{Camera Representation and Injection}
\label{sec:cam_rep_inj}

A central design question is how to encode the target camera trajectory $\{\hat{\mathbf{P}}_i, \mathbf{K}_i\}_{i=0}^{F-1}$ and inject it into the DiT to steer novel-view synthesis. We systematically study two camera representations paired with three injection mechanisms (\cref{fig:camera}), and identify the combination that best supports our unified control framework.

\paragraph{Linear Camera Representation.}
Following ReCamMaster~\cite{bai2025recammaster}, each frame's camera is encoded by flattening the extrinsic matrix $\mathbf{P}_i = [\mathbf{R}_i \mid \mathbf{t}_i] \in \mathbb{R}^{3 \times 4}$ and intrinsic matrix $\mathbf{K}_i \in \mathbb{R}^{3 \times 3}$, then concatenating them into a 21-dimensional vector $\mathbf{c}_i = [\mathrm{vec}(\mathbf{P}_i);\, \mathrm{vec}(\mathbf{K}_i)]$. A learnable linear encoder $\mathcal{E}_\mathrm{cam}: \mathbb{R}^{21} \rightarrow \mathbb{R}^{d}$ projects $\mathbf{c}_i$ to dimension $d$ and broadcasts the embedding to all spatial tokens. This frame-level representation is lightweight but lacks per-pixel geometric detail.

\paragraph{Plücker Ray Representation.}
To obtain per-pixel geometric detail, we adopt Plücker ray embeddings~\cite{sitzmann2021light}, following CameraCtrl~\cite{he2024cameractrl} and BulletTime~\cite{wang2025bullettime}. For each pixel $(u, v)$ in frame $i$, the world-space ray direction and camera origin are:
\begin{equation}
  \mathbf{d}_{i,uv} = \mathbf{R}_i \cdot \hat{\mathbf{d}}_{i,uv}, \quad
  \hat{\mathbf{d}}_{i,uv} = \left[\tfrac{u - c_x}{f_x},\; \tfrac{v - c_y}{f_x},\; 1\right]^\top \big/ \|\cdot\|, \quad \mathbf{o}_i = \mathbf{t}_i.
\end{equation}
The per-pixel embedding $\boldsymbol{\pi}_{i,uv} = [\mathbf{o}_i;\, \mathbf{d}_{i,uv}] \in \mathbb{R}^6$ encodes both viewpoint and focal length at each spatial location, yielding a feature map $\boldsymbol{\Pi} \in \mathbb{R}^{F \times H \times W \times 6}$. After patchification, a learned projection maps these per-token embeddings to the hidden dimension $d$ and makes them available to every DiT block.
This representation also makes resolution editing native to the camera model.
Given a target resolution $H_t{\times}W_t$ and field-of-view, the intrinsic matrix $\mathbf{K}_t$ determines the ray direction for every pixel.
When the resolution changes, the ray directions, principal point, and number of rays change accordingly, so the Plücker ray field naturally adapts to the new imaging geometry instead of relying on cropping, masks, or outpainting.

\paragraph{Token-wise Addition.}
The simplest injection strategy directly adds the projected camera embedding to the hidden state prior to self-attention:
\begin{equation}
  \tilde{\mathbf{x}} = \mathbf{x} + \mathcal{E}_\mathrm{cam}(\mathbf{c}),
\end{equation}
where $\mathbf{c}$ denotes the camera embedding for the current token~\cite{bai2025recammaster,he2024cameractrl}. Although straightforward and zero-initializable, additive injection directly modifies the hidden-state distribution of the pretrained backbone, which can introduce training instability when optimizing multiple control dimensions simultaneously.

\paragraph{AdaLN Injection.}
Adaptive Layer Normalization (AdaLN) instead uses the camera embedding to predict per-token scale and shift parameters~\cite{wang2025bullettime,he2025cameractrl}:
\begin{equation}
  \tilde{\mathbf{x}} = (1 + \boldsymbol{\gamma}) \odot \mathbf{x} + \boldsymbol{\beta},\quad
  [\boldsymbol{\gamma}, \boldsymbol{\beta}] = \mathrm{MLP}\bigl(\mathcal{E}_\mathrm{cam}(\mathbf{c})\bigr),
\end{equation}
where $\boldsymbol{\gamma}$ and $\boldsymbol{\beta}$ are zero-initialized so the injection begins as identity. AdaLN modulates feature distributions without directly overwriting token content, providing smoother incorporation of the camera signal while preserving the pretrained model's generative prior.

\paragraph{RoPE Injection.}
An orthogonal design integrates camera geometry directly into the attention mechanism. ReDirector~\cite{park2025redirector} proposes Rotary Camera Encoding (RoCE), which augments the spatial components of the standard 3D RoPE with learned, Plücker-conditioned phase shifts $\boldsymbol{\phi}$:
\begin{equation}
  \bar{\mathbf{q}}' = \bar{\mathbf{q}} \circ \mathbf{R}_\mathrm{rope} \circ e^{i\boldsymbol{\phi}}, \quad
  \bar{\mathbf{k}}' = \bar{\mathbf{k}} \circ \mathbf{R}_\mathrm{rope} \circ e^{i\boldsymbol{\phi}},
\end{equation}
where $\boldsymbol{\phi}$ is predicted by a lightweight MLP and zero-initialized for stable fine-tuning. Encoding camera information as a phase shift in the complex rotation domain naturally modulates the attention similarity: tokens sharing consistent viewpoints receive boosted attention while attention to geometrically distant tokens is suppressed.
The original 3D RoPE is also resolution-aware: its spatial coordinates are computed from the patchified latent grid $(h',w')$, where $h'{=}H_t/(s_\text{vae}{\cdot}s_\text{patch})$ and $w'{=}W_t/(s_\text{vae}{\cdot}s_\text{patch})$.
Changing the output resolution directly changes this latent grid, allowing RoPE frequencies to adapt to the new spatial layout without architectural changes.

\paragraph{Design Choice.}
Our ablation (Sec.~\ref{sec:ablation}) shows that Plücker rays with RoPE injection achieve the best reconstruction fidelity. Per-pixel Plücker embeddings capture spatially varying camera geometry beyond ReCamMaster's frame-level $\mathbb{R}^{21}$ encoding, and RoPE injection incorporates multi-view relationships without perturbing pretrained features.
At inference time, generating a video at a new target resolution only requires constructing $\mathbf{K}_t$ from the desired $(H_t,W_t,\mathrm{hfov}_t)$, computing per-pixel Plücker rays from $\mathbf{K}_t$ and the target extrinsic poses, and running the DiT on the corresponding latent grid.
No retraining, padding, masks, or post-processing are required.

\subsection{Training Strategy}

We decompose our video-to-video generation task into three orthogonal control dimensions: (1) camera extrinsics (motion/cut), (2) focal length, and (3) aspect ratio. During training, each sample is constructed by independently sampling along these three axes, enabling the model to learn each control dimension as well as their arbitrary combinations.

Specifically, for each training iteration, we first sample the extrinsics type from {same-camera, different single-shot, single-to-multi-shot} with probabilities {0.2, 0.4, 0.4}. The same-camera mode keeps identical camera trajectory between the condition and target videos, while different single-shot draws two distinct single-shot cameras from the same scene, and single-to-multi-shot pairs a single-shot condition with a multi-shot target. We then independently decide whether to apply a focal length change (probability 0.5) and an aspect ratio change (probability 0.2). When focal change is activated, the target video is drawn from a scene rendered with a different focal length while maintaining the same camera extrinsics, allowing the model to disentangle intrinsic and extrinsic camera parameters. For aspect-ratio and resolution editing, the target camera rays and latent grid are constructed at the target output resolution, so the model learns native resolution adaptation rather than fixed-resolution cropping or outpainting. To ensure every training sample carries a meaningful learning signal, we enforce that at least one dimension must differ between the condition and target; if the sampled configuration yields no change (same camera, no focal change, no aspect ratio change), we promote it to a focal-change sample.

This orthogonal decomposition yields up to 3×2×2 = 12 task combinations, covering both single-dimension control (changing only extrinsics, focal length, or aspect ratio) and compound control instructions (e.g., simultaneously changing the camera trajectory, focal length, and aspect ratio).

\section{Experiments}

\subsection{Experiment Settings}

\paragraph{Implementation details.}
Our model is built upon Wan2.1-T2V-1.3B~\cite{wan2025wan}, with only the self-attention layers, the camera encoder, and the camera adapter modules set as trainable; all remaining backbone parameters are kept frozen.
For each training sample, the source video $V_s$ is always drawn from the single-shot, fixed-focal subset of our dataset (Sec.~\ref{sec:dataset}), while the target video $V_t$ is selected according to the sampled task dimensions (extrinsic change, focal change, or resolution change) as described in Sec.~\ref{sec:method}.
The resolution of each sample is drawn from a predefined bucket set \{$480{\times}832$, $832{\times}480$, $384{\times}672$, $672{\times}384$, $512{\times}672$, $672{\times}512$, $384{\times}512$, $512{\times}384$, $640{\times}640$, $320{\times}320$\}; when the resolution-change dimension is active, source and target resolutions are sampled independently to guarantee an aspect-ratio difference.
The newly introduced camera modules are zero-initialized prior to training.
We train for 20k steps with a learning rate of $5{\times}10^{-5}$ and a batch size of 32.

\paragraph{Baselines.}
We evaluate against two groups of methods aligned with the two task categories in our benchmark.
For \textit{extrinsic and focal control}, we compare with
\textbf{TrajectoryCrafter}~\cite{Yu_2025_ICCV}, which reconstructs a point-cloud representation from the input video via monocular depth estimation and renders novel-view frames under target camera poses, filling occluded regions with a video diffusion prior;
and \textbf{ReCamMaster}~\cite{bai2025recammaster}, which conditions a video diffusion model on target camera extrinsic matrices using multi-camera synchronized training data rendered with Unreal Engine~\cite{ue}.
For \textit{resolution editing}, we compare with
\textbf{Follow-Your-Canvas}~\cite{chen2025infinite}, a video outpainting method that expands the spatial extent of a video while preserving temporal coherence;
and \textbf{VACE}~\cite{vace}, a general video editing framework used here as a reference-based generation baseline.

\paragraph{Evaluation metrics.}
We evaluate on both synthetic and real-world videos.
\textbf{\textit{For synthetic evaluation}}, we sample 50 unseen scenes from our rendered dataset and construct 6 test cases per scene, yielding 300 test pairs in total.
The source video for every test case is a fixed-focal, single-shot recording from the same scene.
The six target configurations cover all combinations of extrinsic and intrinsic control dimensions:
(1)~\textit{source portrait}: the source video re-encoded at portrait resolution ($832{\times}480$) as a resolution-change reference;
(2)~\textit{single-shot extrinsics}: a different single-shot camera trajectory at the same focal length;
(3)~\textit{multi-shot extrinsics}: a multi-shot (cut) trajectory at the same focal length;
(4)~\textit{focal only}: the source trajectory rendered at a different focal length;
(5)~\textit{single-shot extrinsics + focal}: a different single-shot trajectory combined with a focal-length change;
(6)~\textit{multi-shot extrinsics + focal}: a multi-shot trajectory combined with a focal-length change.
We report PSNR~$\uparrow$, SSIM~$\uparrow$, and LPIPS~$\downarrow$~\cite{zhang2018unreasonable} to measure reconstruction fidelity against the ground-truth target videos.
Perceptual quality is assessed using FVD~$\downarrow$~\cite{unterthiner2019fvd}, and six VBench~\cite{huang2024vbench} dimensions: aesthetic quality, imaging quality, temporal flickering, motion smoothness, subject consistency, and background consistency.
\textbf{\textit{For real-world evaluation}}, we select 50 videos from the DAVIS dataset~\cite{DBLP:journals/corr/Pont-TusetPCASG17} and pair each with 6 synthesized camera trajectories following the same scheme, yielding 300 test pairs.
Since no ground-truth target videos exist for these pairs, we evaluate \textit{camera accuracy} and \textit{perceptual quality}. For camera accuracy, we run ViPE~\cite{DBLP:journals/corr/abs-2508-10934} on each generated video to estimate per-frame poses and compare against the target trajectory. We report \textbf{RotErr}\,($^{\circ}$, $\downarrow$), the mean rotation error after normalising both trajectories to frame 0, and \textbf{TransErr}\,($\downarrow$), the mean translation error after scale alignment via arc length. Perceptual quality is evaluated with VBench.

\subsection{Comparisons with state-of-the-art}
\begin{table}[t]
  \centering
  \caption{Quantitative comparison with state-of-the-art methods on the synthetic dataset. We report three categories of evaluation: Reconstruction (Recon.), VBench \cite{huang2024vbench}, and Video Quality (V.Q.). Extrinsic and focal length controllability are evaluated against TrajectoryCrafter (Traj.) \cite{Yu_2025_ICCV} and ReCamMaster (ReCam.) \cite{bai2025recammaster}, while resolution adaptation is compared with Follow-Your-Canvas (F-Y-C) \cite{chen2025infinite} and VACE \cite{vace}.}
  \label{tab:quant_synth_all}
  \setlength{\tabcolsep}{3pt}
  \resizebox{0.9\textwidth}{!}{
  \begin{tabular}{ll|ccc|ccc}
    \hline
    \multirow{2}{*}{Category} & \multirow{2}{*}{Metric} & \multicolumn{3}{c|}{Ext./Focal} & \multicolumn{3}{c}{Resolution Editing} \\
    \cline{3-8}
    & & Traj. & ReCam. & \textbf{Ours} & F-Y-C & VACE & \textbf{Ours} \\
    \hline
    \multirow{3}{*}{Recon.}
      & PSNR$\uparrow$    & 12.24  & 12.87  & \textbf{15.88}  & 19.91  & 19.45  & \textbf{20.54} \\
      & SSIM$\uparrow$    & 0.379  & 0.379  & \textbf{0.482}  & \textbf{0.765} & 0.714  & 0.758 \\
      & LPIPS$\downarrow$ & 0.654  & 0.607  & \textbf{0.179}  & 0.206  & 0.194  & \textbf{0.174} \\
    \hline
    \multirow{6}{*}{VBench}
      & Aesthetic$\uparrow$  & 0.4726 & 0.5338 & \textbf{0.5340} & 0.5374 & 0.5460 & \textbf{0.5462} \\
      & Imaging$\uparrow$    & 0.5807 & 0.6505 & \textbf{0.6695} & 0.6346 & 0.6598 & \textbf{0.6614} \\
      & Flicker$\uparrow$    & 0.9565 & \textbf{0.9717} & 0.9648 & \textbf{0.9670} & 0.9652 & 0.9655 \\
      & Motion$\uparrow$     & 0.9845 & \textbf{0.9910} & 0.9881 & 0.9892 & 0.9889 & \textbf{0.9903} \\
      & Subject$\uparrow$    & 0.8619 & \textbf{0.9084} & 0.8912 & 0.8929 & 0.9014 & \textbf{0.9030} \\
      & Background$\uparrow$ & 0.9131 & \textbf{0.9183} & 0.9071 & 0.9133 & \textbf{0.9225} & 0.9100 \\
    \hline
    V.Q.
      & FVD$\downarrow$      & 462.9  & 351.5  & \textbf{231.0}  & 526.8  & 422.7  & \textbf{325.8} \\
    \hline
  \end{tabular}
  }
\end{table}

\paragraph{Quantitative results.}

Quantitative comparisons are summarized in \cref{tab:quant_synth_all,tab:quant_real_all}, grouped into \textit{extrinsic/focal control} and \textit{resolution editing} with task-specific baselines. For \textit{extrinsic and focal control}, camera following is the primary objective, and our method follows the target camera more accurately. It improves synthetic reconstruction over both TrajectoryCrafter~\cite{Yu_2025_ICCV} and ReCamMaster~\cite{bai2025recammaster} (PSNR: \textbf{15.88} vs.\ 12.24 and 12.87) and achieves lower real-world camera error (RotErr: \textbf{2.76}$^\circ$ vs.\ 5.12$^\circ$ for ReCamMaster and 13.99$^\circ$ for TrajectoryCrafter; TransErr: \textbf{0.33} vs.\ 0.46 and 0.99). ReCamMaster scores higher on several VBench dimensions because it keeps the first frame identical to the source and does not perform multi-shot transitions, thereby preserving more input pixels and synthesizing fewer novel regions. In contrast, our method targets harder novel-view and camera-cut settings, yielding substantially stronger camera accuracy and better Imaging Quality on both benchmarks. \Cref{tab:combined_eval} further reports a user study showing higher human preference for camera alignment and video quality.

\begin{table}[t]
  \centering
  \caption{Quantitative comparison on real-world DAVIS videos. We report Camera Accuracy (Cam.\ Acc.) and VBench~\cite{huang2024vbench}. Extrinsic camera control is compared with TrajectoryCrafter (Traj.)~\cite{Yu_2025_ICCV} and ReCamMaster (ReCam.)~\cite{bai2025recammaster}. Resolution adaptation is compared with Follow-Your-Canvas (F-Y-C)~\cite{chen2025infinite} and VACE~\cite{vace}. ``--'' indicates the method does not support this task.}
  \label{tab:quant_real_all}
  \setlength{\tabcolsep}{3pt}
  \resizebox{0.9\textwidth}{!}{
  \begin{tabular}{ll|ccc|ccc}
    \hline
    \multirow{2}{*}{Category} & \multirow{2}{*}{Metric} & \multicolumn{3}{c|}{Ext./Focal} & \multicolumn{3}{c}{Resolution Editing} \\
    \cline{3-8}
    & & Traj. & ReCam. & \textbf{Ours} & F-Y-C & VACE & \textbf{Ours} \\
    \hline
    \multirow{2}{*}{Cam.\ Acc.}
      & RotErr$\downarrow$   &  13.9916 & 5.11982  & \textbf{2.7600}  & --  & --  & -- \\
      & TransErr$\downarrow$ & 0.9949 & 0.4628 & \textbf{0.3309} & --  & --  & -- \\
    \hline
    \multirow{6}{*}{VBench}
      & Aesthetic$\uparrow$  & 0.4291 & \textbf{0.5166} & 0.4752 & 0.5166 & 0.5157 & \textbf{0.5198} \\
      & Imaging$\uparrow$    & 0.6004 & 0.6833 & \textbf{0.6857} & 0.6728 & 0.6644 & \textbf{0.6732} \\
      & Flicker$\uparrow$    & 0.9077 & \textbf{0.9596} & 0.9463 & 0.9408 & 0.9395 & \textbf{0.9413} \\
      & Motion$\uparrow$     & 0.9527 & \textbf{0.9884} & 0.9809 & 0.9718 & 0.9743 & \textbf{0.9790} \\
      & Subject$\uparrow$    & 0.8142 & \textbf{0.9094} & 0.8410 & 0.8898 & \textbf{0.8925} & 0.8908 \\
      & Background$\uparrow$ & 0.8857 & \textbf{0.9110} & 0.8828 & 0.9125 & \textbf{0.9271} & 0.9127 \\
    \hline
  \end{tabular}
  }
\end{table}

For \textit{resolution editing}, our method performs consistently well across synthetic and real-world evaluations, with advantages in reconstruction (PSNR/LPIPS), video quality (FVD), and key perceptual metrics such as Aesthetic, Imaging Quality, and Motion Smoothness over Follow-Your-Canvas~\cite{chen2025infinite} and VACE~\cite{vace}.
VACE attains slightly higher Subject and Background Consistency because its outpainting pipeline preserves the source crop, while our method re-generates the full frame to support native resolution adaptation.
These results show the benefit of jointly modeling camera geometry and resolution changes within a unified framework.

\begin{figure}[t]
  \centering
  \includegraphics[width=\linewidth]{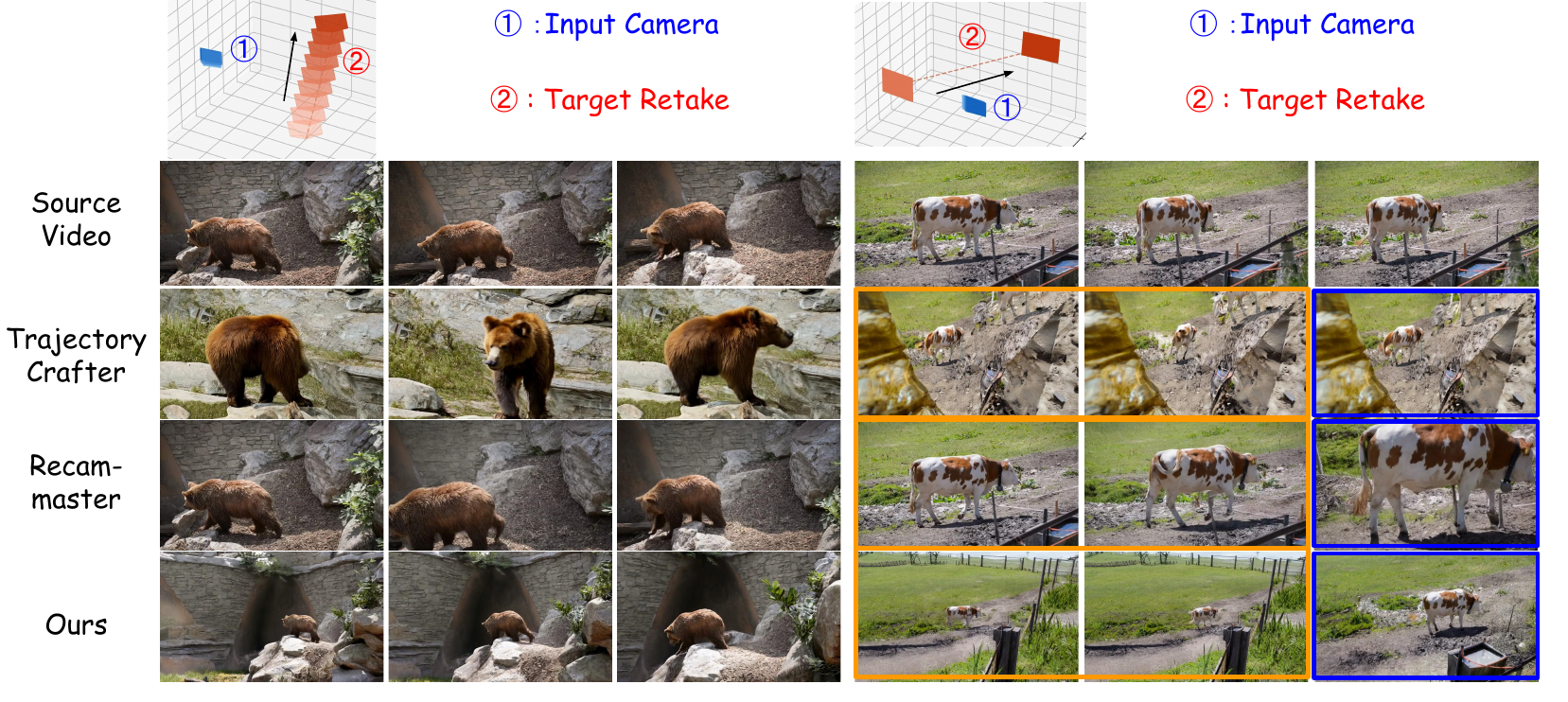}
  \caption{Qualitative comparison on real-world videos with TrajectoryCrafter~\cite{Yu_2025_ICCV} and ReCamMaster~\cite{bai2025recammaster}. Given the same source video and target retake camera parameters, our method demonstrates more accurate camera trajectory following under challenging scenarios, including large viewpoint changes and multi-cut shot transitions, while maintaining higher visual fidelity and geometric consistency.}
  \label{fig:qualitative}
\end{figure}

\begin{figure}[t]
  \centering
  \includegraphics[width=\linewidth]{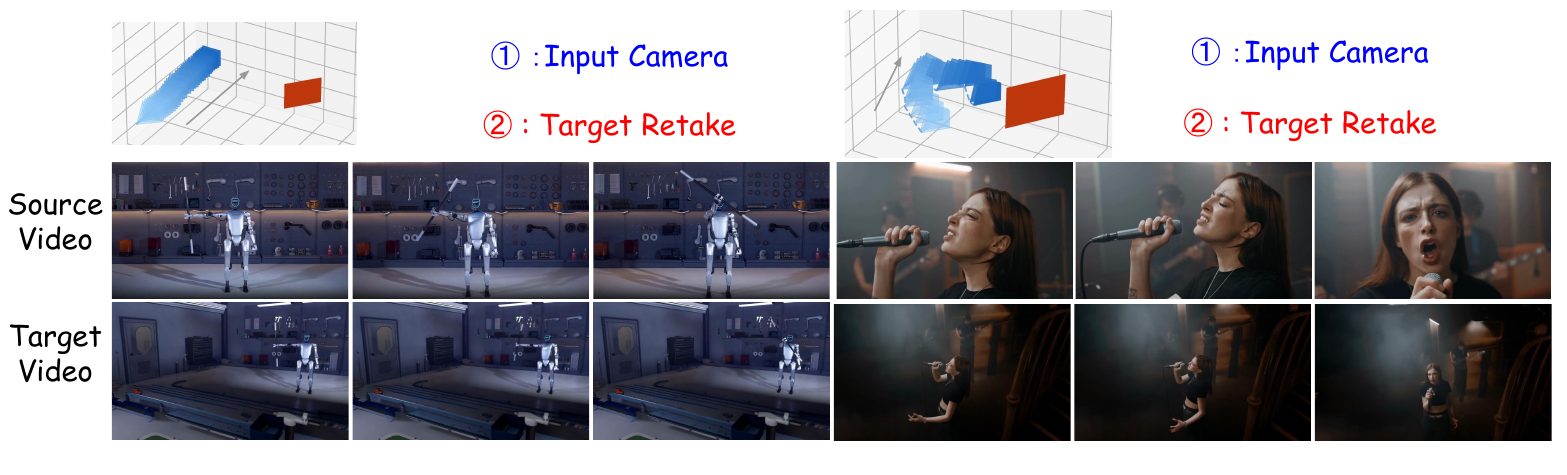}
  \caption{In-the-wild results. Our method generalizes to diverse real-world scenes despite being trained on synthetic data only.}
  \label{fig:more_results}
\end{figure}

\paragraph{Qualitative results.}
\cref{fig:qualitative} shows real-world comparisons under large viewpoint changes and multi-shot retake trajectories.
TrajectoryCrafter often suffers from geometric errors and visual artifacts, while ReCamMaster~\cite{bai2025recammaster} keeps the first frame identical to the source video and struggles to follow multi-shot camera transitions.
Our method follows complex camera controls more accurately while preserving geometric consistency and visual quality. The in-the-wild examples in \cref{fig:more_results} further demonstrate its generalization to real scenes despite training only on synthetic data.

\begin{table}[t]
  \centering
  \caption{Camera representation and injection ablation. ``Linear'' denotes a frame-level 21-d encoding of extrinsic and intrinsic matrices; ``Pl\"ucker Ray'' denotes per-pixel 6-DoF ray embeddings~\cite{sitzmann2021light}. Injection methods are element-wise addition (Add), adaptive layer normalization (AdaLN), and RoPE-based phase shift (RoPE).}
  \label{tab:ablation_cam}
  \setlength{\tabcolsep}{10pt}
  \renewcommand{\arraystretch}{1.15}
  \begin{tabular}{ll ccc}
    \toprule
    Representation & Injection & PSNR$\uparrow$ & SSIM$\uparrow$ & LPIPS$\downarrow$ \\
    \midrule
    Linear & Addition & 15.62 & 0.481 & 0.434 \\
    Pl\"ucker Ray  & AdaLN     & 16.51 & 0.517 & 0.368 \\
    Pl\"ucker Ray  & RoPE      & \textbf{16.66} & \textbf{0.528} & \textbf{0.372} \\
    \bottomrule
  \end{tabular}
\end{table}

\begin{figure*}[t]
  \centering
  \includegraphics[width=\linewidth]{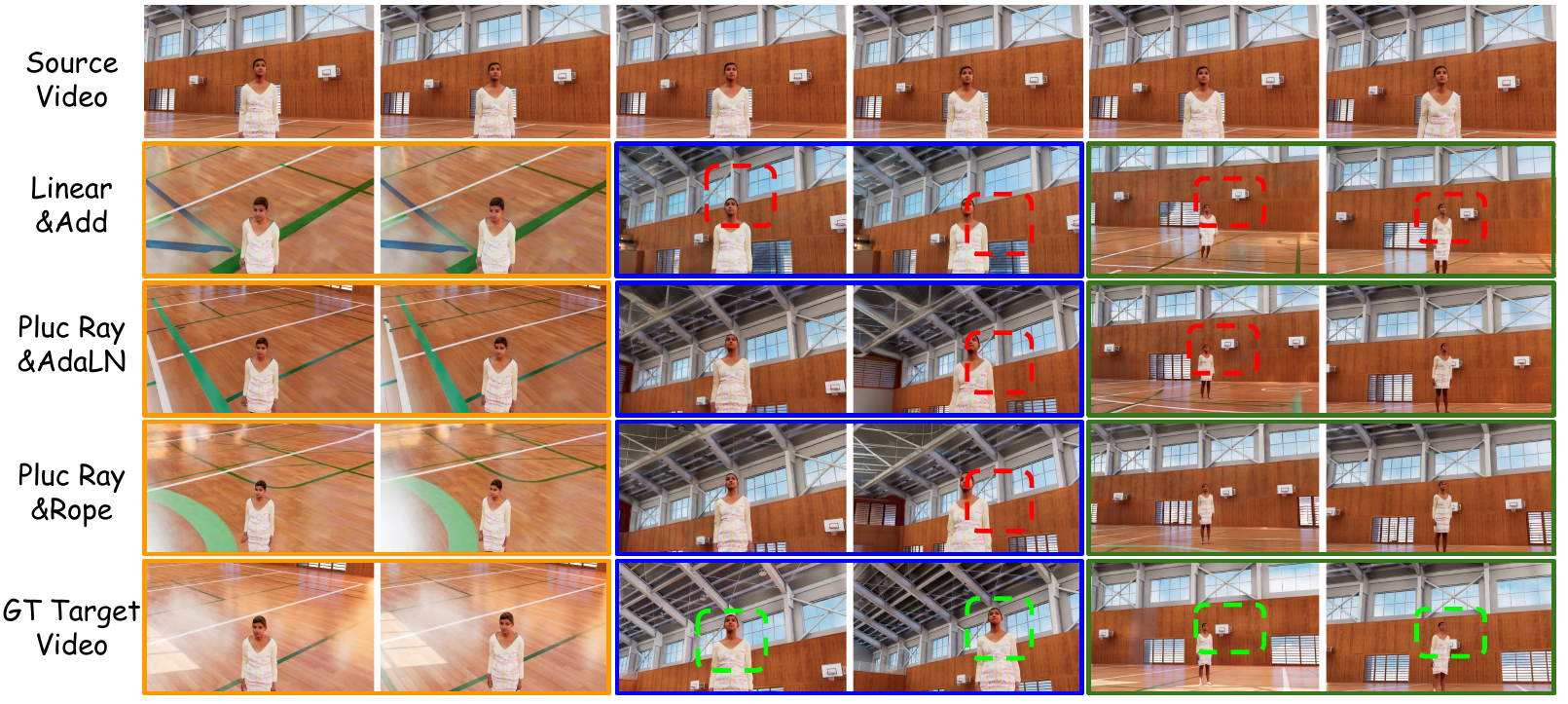}
  \caption{Visual comparisons for the ablation study. Given the same source video and target retake camera parameters, Pl\"ucker ray representation with RoPE-based injection yields stronger geometric consistency, with fewer artifacts (highlighted by red dashed boxes) compared to other variants. Green dashed boxes indicate the ground truth.}
  \label{fig:ablation}
\end{figure*}

\subsection{Ablation studies and analysis}
\label{sec:ablation}
\begin{table}[t]
  \centering
  \caption{Training strategy ablation on the synthetic benchmark. \textit{Extrinsic-only} disables focal/resolution changes; \textit{W/o identity samples} makes all control dimensions change together. Best in \textbf{bold}.}
  \label{tab:ablation_train_results}
  \setlength{\tabcolsep}{3pt}
  \renewcommand{\arraystretch}{1.15}
  \resizebox{\linewidth}{!}{
  \begin{tabular}{l ccc}
    \toprule
    \multirow{2}{*}{Configuration} & Extrinsic Tasks & Focal Tasks & Resolution Tasks \\
    \cmidrule(lr){2-2}\cmidrule(lr){3-3}\cmidrule(lr){4-4}
    & PSNR$\uparrow$ / SSIM$\uparrow$ / LPIPS$\downarrow$ & PSNR$\uparrow$ / SSIM$\uparrow$ / LPIPS$\downarrow$ & PSNR$\uparrow$ / SSIM$\uparrow$ / LPIPS$\downarrow$ \\
    \midrule
    Full model
      & \textbf{14.13} / \textbf{0.423} / \textbf{0.483}
      & \textbf{20.03} / \textbf{0.606} / \textbf{0.213}
      & \textbf{20.05} / \textbf{0.743} / \textbf{0.187} \\
    Extrinsic-only
      & 14.00 / 0.414 / 0.497
      & -- & -- \\
    W/o identity samples
      & 13.87 / 0.406 / 0.503
      & 15.94 / 0.475 / 0.399
      & 14.64 / 0.456 / 0.414 \\
    \bottomrule
  \end{tabular}
  }
\end{table}
\paragraph{Camera representation and injection.}
Table~\ref{tab:ablation_cam} compares three representation and injection choices with all other settings fixed.
Compared with linear representation and token-wise addition (row 1), row 2 improves all metrics by using Pl\"ucker rays for per-pixel geometric cues and AdaLN to modulate token activations through learned scale and shift, instead of directly perturbing their values.
Replacing AdaLN with RoPE injection (row 3) further improves performance, suggesting that injecting camera geometry through attention is more effective than feature-space modulation. We include qualitative comparisons for different injection designs in \cref{fig:ablation}.

\paragraph{Training strategy.}
Table~\ref{tab:ablation_train_results} ablates our joint training strategy with two variants: \textit{extrinsic-only} and \textit{without identity samples}.
Extrinsic-only training disables focal and resolution changes, yet performs comparably to the full model on extrinsic tasks, confirming that jointly training focal length and resolution does not degrade extrinsic control.
Without identity samples, all control dimensions change together, and performance drops consistently across all three task groups.
This shows that identity samples provide the data-driven supervision needed to decouple extrinsic, focal, and resolution controls during joint training.

\section{Conclusion}

We presented CameraAnything, a unified framework for camera-controlled video editing that supports multi-shot retakes, focal-length, and resolution changes in a single generation process.
We inject per-pixel Pl\"ucker rays into resolution-aware 3D RoPE in self-attention.
We further built a scalable synthetic dataset via structured multi-camera recording, with synchronized clip pairs covering extrinsic, focal, and resolution variations.
Moreover, we designed an orthogonal training strategy that independently samples extrinsic, focal, and resolution edits, enabling both single-dimension control and their compound combinations.
Experiments demonstrate strong benchmark performance and practical promise for cinematic production and cross-platform video adaptation.

\section*{Acknowledgements}
This work was supported by Ant Group Research Intern Program and Ant Group Postdoctoral Programme.

\bibliographystyle{splncs04}
\bibliography{main}

\appendix
\clearpage
\begin{center}
  {\LARGE\bfseries CameraAnything: Refilming Videos with Arbitrary Camera Control\par}
  \vspace{0.8em}
  {\large Supplementary Material\par}
\end{center}
\vspace{1.5em}
\section{Dataset Details}
\label{sec:supp_dataset}

\paragraph{Scene construction.}
Our dataset is rendered using Unreal Engine 5~\cite{ue} and comprises a total of \textbf{15,000} scenes built from \textbf{150} static environments spanning diverse indoor and outdoor settings.
For each scene, we randomly place animated human characters from the BEDLAM~\cite{DBLP:conf/cvpr/BlackPTY23} dataset into the environment and generate camera trajectories conditioned on character positions.
Unreal Engine's built-in collision detection is used throughout the pipeline to ensure that characters, scene geometry, and cameras do not interpenetrate.
Each scene contains \textbf{20} synchronized video clips of \textbf{81} frames at \textbf{16}\,fps and a resolution of \textbf{832$\times$832}. Our dataset have about \textbf{300,000} videos.

\paragraph{Dataset structure.}
The 20 clips per scene are organised into three groups:
\begin{itemize}
    \item \textbf{Group~A} (Clips 1--5): Single-shot videos with smooth, continuous camera trajectories and a fixed focal length. These clips can serve as either source or target during training.
    \item \textbf{Group~B} (Clips 6--10): Multi-shot videos with \textbf{1--3} discrete camera transitions per clip, simulating cinematic shot cuts.
    \item \textbf{Group~C} (Clips 11--20): Re-focal variants of Groups~A and~B with unchanged extrinsic trajectories and editing structure (see below).
\end{itemize}
For Groups~A and~B, each scene uses a single fixed focal length randomly chosen from $\{18, 24, 35, 50\}$\,mm equivalents, following the camera configuration of ReCamMaster~\cite{bai2025recammaster}.

\paragraph{Camera trajectory design.}
We follow the taxonomy proposed in CameraBench~\cite{DBLP:journals/corr/abs-2504-15376} and decompose each shot's trajectory into one of two categories:
\begin{itemize}
    \item \textbf{Camera-centric}: The camera moves independently of scene content. This category includes three subtypes: (i)~\textit{static} (fixed position and orientation), (ii)~\textit{rotation-only} (rotating in place without translation), and (iii)~\textit{translation-only} (translating without rotation change).
    \item \textbf{Object-centric}: The camera orbits around or tracks a target character. At the start and end frames of each shot, we randomly sample a camera-to-subject distance in $[1, 7]$\,m and apply yaw and pitch offsets both drawn from $\mathcal{N}(0^\circ, 15^\circ)$ relative to the subject's centre-facing direction, with the camera always oriented towards the subject. The two endpoint poses are connected via either linear interpolation or a non-uniform arc-curve interpolation.
\end{itemize}
For Group~B, \textbf{2--4} such shots are concatenated with instantaneous camera cuts to form multi-shot sequences.

\paragraph{Re-focal variants.}
Group~C pairs each clip in Groups~A and~B with a re-focal counterpart that preserves extrinsic motion and shot structure while varying intrinsics.
Focal lengths span the 18--50\,mm equivalent range, with values sampled uniformly in FOV space.
Re-focal editing follows several patterns; for multi-shot clips, one pattern is assigned independently to each shot with equal probability:
\begin{itemize}
    \item \textbf{Fixed}: constant focal length throughout a shot.
    \item \textbf{Zoom-in}: focal length increases over the shot.
    \item \textbf{Zoom-out}: focal length decreases over the shot.
    \item \textbf{Breathe}: focal length increases and then decreases within a shot.
    \item \textbf{Random}: 2--4 focal keyframes within a shot, each assigned an independently sampled focal length with interpolation between keyframes.
\end{itemize}
These paired clips support learning re-focal effects and disentangling optical zoom from camera motion.

\section{Extended Evaluation and User Study}
\label[appendix]{sec:supp_extended_eval}

This section complements the synthetic and DAVIS evaluations in the main paper (Tables~1--2), providing additional analyses on in-the-wild camera accuracy, complementary synthetic metrics, human preference, and per-configuration reconstruction metrics.

\paragraph{In-the-wild evaluation.}
We evaluate on \textbf{15 additional in-the-wild videos} with ViPE-estimated source cameras and ReCamMaster run under its official checkpoint and recommended inference settings for a fair comparison.
Except for the DAVIS baseline comparisons, the video demos on our project page are likewise generated from noisy estimated source cameras rather than ground-truth poses.
Table~\ref{tab:combined_eval} shows that CameraAnything achieves lower RotErr/TransErr than ReCamMaster on these videos, consistent with the DAVIS results in the main paper (Table~2).

\paragraph{Synthetic camera accuracy.}
Table~\ref{tab:combined_eval} additionally reports synthetic RotErr/TransErr on the same benchmark as Table~1 in the main paper, listed alongside the in-the-wild and user-study results for reference.
CameraAnything achieves the lowest camera error among the compared methods.

\paragraph{User study.}
Automatic metrics may not fully reflect human perception of camera control.
We conduct a user study with \textbf{20 participants} on \textbf{15 real-world videos} under extrinsic/focal editing.
Each participant views the source video, a target camera-pose visualisation, and anonymised results from competing methods, and selects the preferred output for source consistency, camera alignment, and overall quality.
As summarised in Table~\ref{tab:combined_eval}, CameraAnything is preferred over ReCamMaster on all three criteria, with the largest margin on camera alignment (73.3\% vs.\ 21.0\%).

\begin{table}[t]
  \centering
  \caption{Combined evaluation on in-the-wild videos, synthetic camera accuracy, and a real-world user study. RotErr/TransErr are reported for in-the-wild and synthetic settings; user-study entries denote preference rates (\%).}
  \label{tab:combined_eval}
  \setlength{\tabcolsep}{3pt}
  \renewcommand{\arraystretch}{1.1}
  \resizebox{\linewidth}{!}{
  \begin{tabular}{l|cc|cc|ccc}
    \toprule
    & \multicolumn{2}{c|}{In-the-wild} & \multicolumn{2}{c|}{Synthetic} & \multicolumn{3}{c}{User Study} \\
    Method & RotErr$\downarrow$ & TransErr$\downarrow$ & RotErr$\downarrow$ & TransErr$\downarrow$ & Src.$\uparrow$ & Cam.$\uparrow$ & Qual.$\uparrow$ \\
    \midrule
    Traj. & -- & -- & 7.771 & 0.492 & 1.9 & 5.7 & 2.9 \\
    ReCam. & 5.110 & 0.537 & 3.969 & 0.260 & 41.0 & 21.0 & 36.2 \\
    \textbf{Ours} & \textbf{3.015} & \textbf{0.259} & \textbf{2.123} & \textbf{0.178} & \textbf{57.1} & \textbf{73.3} & \textbf{61.0} \\
    \bottomrule
  \end{tabular}
  }
\end{table}

\paragraph{Per-configuration breakdown.}
Table~\ref{tab:config_breakdown} reports reconstruction metrics across the six test configurations defined in the main paper.
Focal-only and source-portrait edits are easier, whereas extrinsic and extrinsic+focal edits are harder due to the need for larger novel-region synthesis.

\begin{table}[t]
  \centering
  \caption{Per-configuration reconstruction breakdown on the synthetic benchmark.}
  \label{tab:config_breakdown}
  \setlength{\tabcolsep}{3pt}
  \renewcommand{\arraystretch}{1.1}
  \resizebox{\linewidth}{!}{
  \begin{tabular}{lcccccc}
    \toprule
    Metric & Focal Only & Src.\ Portr. & Single Ext. & Multi Ext. & Single Ext.+F & Multi Ext.+F \\
    \midrule
    PSNR$\uparrow$ & 21.78 & 20.54 & 14.67 & 14.49 & 13.98 & 14.49 \\
    SSIM$\uparrow$ & 0.680 & 0.758 & 0.447 & 0.428 & 0.416 & 0.439 \\
    LPIPS$\downarrow$ & 0.178 & 0.174 & 0.448 & 0.460 & 0.512 & 0.465 \\
    \bottomrule
  \end{tabular}
  }
\end{table}

\section{More Qualitative Results}
\label[appendix]{sec:supp_qualitative}

\begin{figure}[t]
  \centering
  \includegraphics[width=\linewidth]{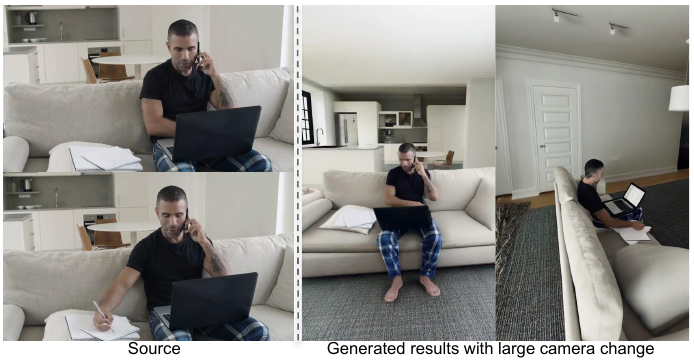}
  \caption{Results under large viewing-angle and resolution changes. Each row shows a source video (left) and our result (right).}
  \label{fig:supp_big_angle}
\end{figure}

\paragraph{Large viewing-angle changes.}
Figure~\ref{fig:supp_big_angle} shows representative in-the-wild results with large changes in camera distance, viewing angle, and resolution from the source video.
Despite the need to synthesise extensive unseen regions, the outputs remain visually coherent and do not exhibit obvious synthetic-looking artifacts.

\begin{figure}[t]
  \centering
  \includegraphics[width=\linewidth]{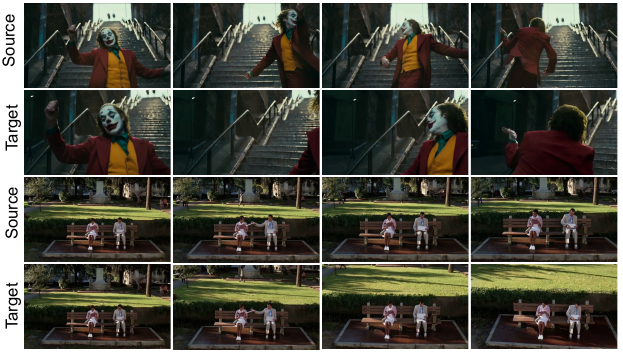}
  \caption{Results on two movie cases. Each row shows a source frame (left) and our edited result (right), illustrating generalization to cinematic footage beyond synthetic training data.}
  \label{fig:supp_film}
\end{figure}

\paragraph{Film clips.}
Following ReCamMaster~\cite{bai2025recammaster}, Figure~\ref{fig:supp_film} shows results on two movie cases, demonstrating the generalization of our method to cinematic content.

\end{document}